\documentclass[onecolumn]{openhelix}

\usepackage{amsmath,amsfonts,bm}

\def\eqref#1{equation~\ref{#1}}

\def\1{\bm{1}}

\DeclareMathAlphabet{\mathsfit}{\encodingdefault}{\sfdefault}{m}{sl}
\SetMathAlphabet{\mathsfit}{bold}{\encodingdefault}{\sfdefault}{bx}{n}

\usepackage{amsmath}
\usepackage{amssymb}
\usepackage{booktabs}
\usepackage{graphicx}
\usepackage{subcaption}
\usepackage{wrapfig}
\usepackage{multirow}
\usepackage{tabularx}
\usepackage{makecell}
\usepackage{url}
\usepackage{float}
\usepackage{hyperref}

\newcommand{\method}{Discrete Forcing}
\title{Discrete Forcing: Infusing Discrete Guidance into Continuous Denoising for Few-Step Action Experts}
\author[1,2,*]{Jingbo Wang}
\author[1,*,\dagger]{Wenxuan Song}
\author[3,*]{Wenhao Yu}
\author[4,5,*]{Han Zhao}
\author[6]{ Xi Wang}
\author[1]{Jiayi Chen}
\author[4]{Donglin Wang}
\author[6]{Yan Wang}
\author[1,\ddagger]{Haoang Li}

\affiliation[1]{The Hong Kong University of Science and Technology (Guangzhou)}
\affiliation[2]{South China University of Technology}
\affiliation[3]{University of Science and Technology of China}
\affiliation[4]{Westlake University}
\affiliation[5]{Zhejiang University}
\affiliation[6]{Tsinghua University}

\contribution[*]{Equal Contribution}
\contribution[\dagger]{Project Lead}
\contribution[\ddagger]{Corresponding Author}

\metadata[Project Page]{\raisebox{-1.5pt}{\includegraphics[height=1.05em]{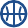}}~\href{https://discrete-forcing.github.io/}{\texttt{discrete-forcing.github.io}}}
\abstract{
Efficient action generation in vision-language-action (VLA) models requires capturing both coarse action structure and fine-grained details.
Discrete action tokens provide compact structural representations but sacrifice precision, while continuous action tokens offer high precision but often require multiple denoising steps.
We introduce Discrete Forcing, a flow-matching framework that combines these representations through an explicit coarse-to-fine generation process. 
It first predicts discrete action tokens to establish a coarse action structure, then uses them to guide continuous action refinement. The discrete and continuous components share a common diffusion transformer backbone with specialized branches, maintaining a parameter count comparable to a conventional single-branch model while requiring only one forward pass per branch.
Extensive evaluations across multiple benchmarks demonstrate improved performance and faster inference over a parameter-matched continuous action expert, with consistent performance gains as model capacity increases. Real-world experiments further demonstrate improvements on high-precision and dynamic manipulation tasks.
}

\begin{document}

\maketitle

\begin{figure}[h]
    \centering
    \includegraphics[width=\linewidth]{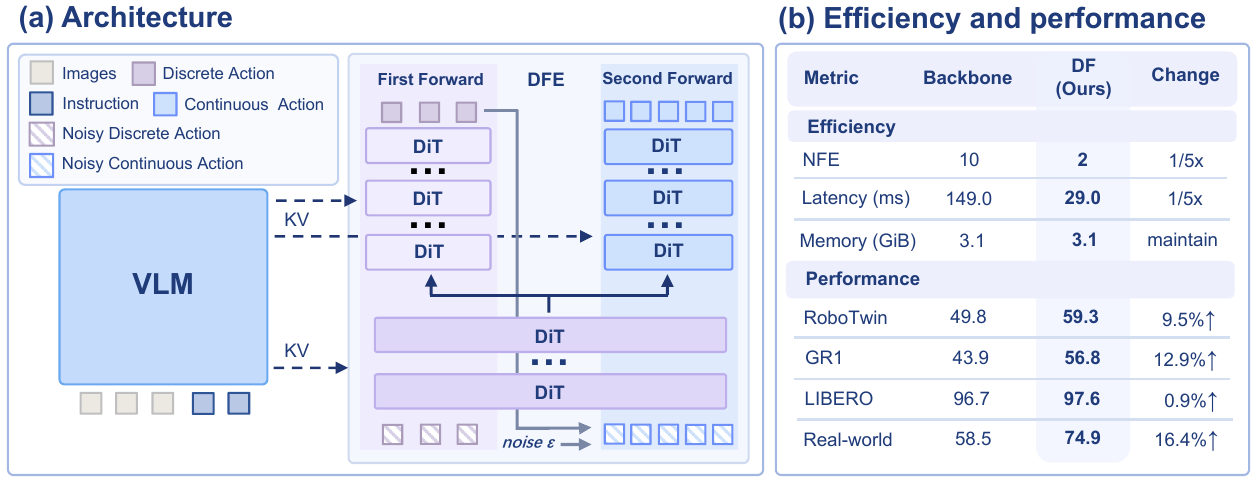}
    \caption{
    \textbf{Overview of the Discrete Forcing (DF).}
(a) Discrete Forcing Expert (DFE) performs discrete action prediction followed by continuous refinement within a partially shared Action DiT conditioned on VLM features.
(b) This two-stage design reduces inference cost while consistently improving performance across simulation and real-world benchmarks.}
    \label{fig:method}
    \vspace{-1em}
\end{figure}

\section{Introduction}
\label{sec:introduction}
Across a wide range of domains, effective generation often proceeds from coarse structure to fine detail: architects establish the layout of a building before construction begins, just as filmmakers write the scenes before shooting~\citep{baade2026latent}. 
In latent diffusion~\citep{ho2020denoising}, this coarse-to-fine progression emerges naturally from the denoising dynamics, shaped by the frequency distribution of natural images.
Thus, pixel-space diffusion invariably means predicting low-frequency details before high-frequency details~\citep{lee2025beta}. In contrast, actions lack a clearly defined frequency-domain structure.
We pose the question: motivated by the coarse-to-fine generation principle observed in other domains, can action generation be modeled in a similar manner?

\begin{figure}[t]
    \centering
    \includegraphics[width=\linewidth]{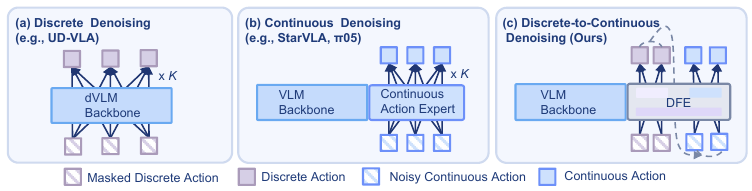}
    \caption{\textbf{Comparison of action generation paradigms.}
Existing methods rely on $K$-step iterative denoising in either (a) discrete or (b) continuous action spaces, whereas (c) \method~(Ours) first predicts a coarse discrete action and then performs a single continuous refinement.}
    \label{fig:compare}
    \vspace{-1.5em}
\end{figure}

\looseness=-1
To address this question, we examine the action generation process from two perspectives in \Cref{fig:compare}: the representation space and the decoding order.
\textbf{1)} In terms of representation, existing action generation methods operate in either discrete or continuous spaces. Discrete action tokens are typically produced by pretrained~\citep{song2026fast, liu2026oat, dong2026actioncodec,chen2026unified} or rule-based tokenizers~\citep{brohan2023rt}; they provide compact representations that capture high-level action information, but often at the cost of reduced precision. Continuous actions, by contrast, offer higher precision and have become the predominant output representation in contemporary Vision-language-action models (VLAs)~\citep{pi05, bjorck2025gr00t, wang2026vla, team2026xiaomi,li2026spatial,zhao2026frappe,bai2026embodied,lei2026robomemarena} and world-action models (WAMs)~\citep{li2026causal, bi2026motus, yang20264d}. However, directly denoising in the continuous action space is at odds with the coarse-to-fine generation principle and typically requires many denoising steps, resulting in slow inference. 
\textbf{2)} Beyond the representation space, the action generation order itself constitutes a broad design space: discrete-action models have explored autoregressive~\citep{kim2024openvla}, block-wise~\citep{song2026fast}, and parallel~\citep{song2025pd} decoding, whereas continuous-action models generally employ either parallel~\citep{chi2025diffusion, wen2024diffusion, wen2025tinyvla,song2026reconvla} or coarse-to-fine generation~\citep{gong2025carp}.

Together, we seek to establish a sequential relationship across different action representations: first denoising discrete tokens that encode high-level information, and then conditioning the subsequent fine-grained denoising of continuous actions on these tokens.
We propose \method, as shown in \Cref{fig:method} (a). In \method, we train a single flow-matching action expert over a discrete space and continuous space simultaneously with multiple time variables. 
The action expert first performs denoising in the discrete space to model the coarse structure of the action.
Then, the generated discrete action effectively serves as a “scratchpad” and is used to initialize the continuous denoising process, thus removing the burden of modeling high-level information from the continuous branch.
To unify the action generation process within a single architecture while explicitly distinguishing the modeling of discrete and continuous spaces, we employ several shared Diffusion Transformer (DiT)~\citep{peebles2023scalable} layers at the beginning of the action expert, followed by two separate branches for discrete and continuous modeling in its intermediate and later stages. The resulting architecture retains the same parameter count as a conventional single-branch DiT. Our experiments demonstrate that this design achieves the best performance.
Benefiting from its explicit coarse-to-fine denoising process, Discrete Forcing enables highly efficient action modeling: high-quality actions can be generated with only one forward pass through each of the discrete and continuous branches, substantially improving inference efficiency.

We conduct extensive experiments across multiple benchmarks, demonstrating that our \method ~achieves superior performance and faster inference than a conventional continuous action expert with parallel parameters (\Cref{fig:method} (b)). 
We further validate the scalability of \method~ across models of different sizes, showing that its performance improves consistently with increasing model capacity. 
Real-world experiments further demonstrate that our \method yields performance gains on both high-precision and dynamic manipulation tasks.
In summary, our contributions are as follows:
\begin{itemize}
    \item 
    We propose Discrete Forcing, a flow-matching framework that generates actions through discrete-to-continuous denoising, from coarse structure to precise details.
    \item We design a unified action expert with shared layers and specialized branches, enabling action generation with one forward pass per branch while matching the parameter count of a conventional DiT.
    \item Extensive benchmark and real-world experiments demonstrate superior performance, faster inference, and consistent improvements with increasing model capacity, including gains on high-precision and dynamic manipulation tasks.
\end{itemize}
\vspace{-4mm}
\section{Related Work}
\label{sec:related-work}

\paragraph{Action Representations in VLA Models.}
Action representations shape the learning objectives and decoding mechanisms of VLA policies.
OpenVLA~\citep{kim2024openvla} predicts quantized action tokens autoregressively; Discrete Diffusion VLA~\citep{liang2025discrete} and DFM-VLA~\citep{chen2026dfmvla} instead reconstruct and refine token sequences through discrete denoising and probability velocity fields, respectively.
UD-VLA~\citep{chen2025udvla} extends discrete denoising to joint future-image and action generation.
Continuous action modeling has developed through Diffusion Policy~\citep{chi2025diffusion}, RDT-1B~\citep{liu2024rdt}, and diffusion-based VLAs such as TinyVLA and Diffusion-VLA~\citep{wen2025tinyvla,wen2024diffusion}, with $\pi_0$ and $\pi_{0.5}$~\citep{black2024pi0,pi05} adopting flow-matching action experts.
Hybrid formulations combine both representations: HybridVLA~\citep{liu2025hybridvla} jointly trains autoregressive and diffusion-based action prediction and adaptively ensembles their outputs, while Knowledge Insulation~\citep{driess2025knowledge} trains the backbone with discrete action supervision and blocks gradients from a continuous action expert to preserve knowledge transfer.
These approaches establish complementary roles for discrete and continuous modeling, motivating an explicit study of how the two predictions should interact during action generation.

\paragraph{Coarse-to-Fine Action Generation.}
Coarse-to-fine policies organize action generation around intermediate predictions that guide subsequent refinement.
CARP~\citep{gong2025carp} learns multi-scale quantized action representations and generates them through coarse-to-fine next-scale prediction.
Libra-VLA~\citep{wei2026libra} introduces a discrete--continuous hierarchy in which coarse intent embeddings condition a separate diffusion refiner; an asynchronous intent buffer decouples the frequencies of planning and control.
CF-VLA~\citep{cfvla} constructs an action-aware initialization by learning a continuous endpoint-velocity distribution, followed by a single local refinement step.
Unlike prior hierarchical policies that use discrete representations as semantic intents or planning variables, \method~constructs the coarse and fine stages as two representations of the same action trajectory, allowing the discrete prediction to directly define the source distribution of continuous flow matching.

\paragraph{Few-Step and Efficient Action Generation.}
\looseness=-1
Reducing action-generation cost has motivated parallel decoding, policy distillation, and training objectives tailored to limited inference budgets.
PD-VLA~\citep{song2025pd} combines action chunking with parallel decoding, while OpenVLA-OFT~\citep{kim2025oft} integrates parallel prediction, continuous outputs, and $L_1$ regression.
Consistency Policy~\citep{prasad2024consistency} enforces consistency along a pretrained diffusion policy's trajectories, and OneDP~\citep{wang2025onedp} distills a single-step generator through distribution matching to a diffusion teacher.
For token-based VLAs, CEED-VLA~\citep{song2025ceed} combines consistency distillation with early-exit decoding, and Fast-dVLA~\citep{song2026fast} uses asymmetric distillation and pipelined block-parallel decoding.
MIP~\citep{mip} further shows that iterative two-step regression with intermediate supervision and controlled stochasticity can achieve strong control performance.
 Unlike these approaches that accelerate existing generation paradigms through parallelization or distillation, \method~introduces a discrete-to-continuous generation process with one forward evaluation for each stage, directly optimized from action demonstrations.

\section{Method}
\label{sec:method}

\subsection{Preliminary: Denoising-based Action Generation}
Diffusion-based policies formulate continuous action generation as an iterative denoising process. 
Given an action chunk $\mathbf{A}_t = [\mathbf{a}_t, \ldots, \mathbf{a}_{t+H-1}]$, 
a noisy action $\mathbf{A}_t^\tau$ is constructed by interpolating the clean action with Gaussian noise:
\vspace{-1mm}
\begin{equation}
    \mathbf{A}_t^\tau
    = \tau \boldsymbol{\epsilon} + (1-\tau)\mathbf{A}_t,
    \qquad
    \boldsymbol{\epsilon} \sim \mathcal{N}(\mathbf{0}, \mathbf{I}),
\end{equation}
where $\tau \in [0,1]$ denotes the noise level. 
A conditional denoising network $f_\theta$ takes the noisy action together with the observation $\mathbf{o}_t$ and predicts the conditional vector field associated with this interpolation path. 
Following the flow-matching formulation, the model can be trained with
\begin{equation}
    \mathcal{L}_{\mathrm{FM}}
    =
    \mathbb{E}_{\mathbf{A}_t,\boldsymbol{\epsilon},\tau}
    \left[
    \left\|
    f_\theta(\mathbf{A}_t^\tau, \mathbf{o}_t)
    - (\boldsymbol{\epsilon}-\mathbf{A}_t)
    \right\|_2^2
    \right].
\end{equation}
At inference time, action generation starts from Gaussian noise and progressively integrates the predicted vector field from $\tau=1$ to $\tau=0$, yielding a continuous action chunk conditioned on the current observation.
\subsection{Architecture of Discrete Forcing Head}

As illustrated in Fig.~\ref{fig:method} (a), \method~integrates a VLM with a \method~Expert built upon a partially shared Action DiT. The expert combines shared Transformer layers with specialized discrete and continuous branches, conditioned on layer-wise visual-language features from the VLM through cross-attention. This design supports coarse-to-fine action generation, in which discrete predictions capture the initial action structure and guide subsequent continuous refinement.

\textbf{Partially shared dual-branch Action DiT.}
We construct a partially shared Action DiT with a discrete branch and a continuous branch. The first $L_s$ Transformer layers share parameters across the two action representations, enabling the model to learn common action features. The subsequent layers and output heads are parameterized independently for discrete prediction and continuous refinement.
This partially shared design allows each branch to generate its output with a single forward pass while maintaining the same parameter count as a conventional Action DiT, introducing no additional model-size overhead.

\textbf{Discrete and continuous action representations.}
To construct the input representations, we first normalize the action chunk $\mathbf{A}_t$. For the discrete representation, each scalar entry is independently quantized into one of 255 action bins. Together with a special \texttt{[MASK]} token representing unknown action entries, these bins form a vocabulary of size $K=256$. The resulting indices are mapped to discrete action tokens through a learnable embedding table. For the continuous representation, the normalized action chunk is projected into a sequence of continuous action tokens through an MLP-based action encoder. Both representations use the same hidden dimension before entering the Discrete Forcing Expert.

Within the Discrete Forcing Expert, interleaved self-attention and cross-attention layers model interactions among action tokens and incorporate VLM context, respectively. At each cross-attention layer, the corresponding layer-wise VLM representation provides visual and linguistic conditioning to both action representations throughout generation. The specialized branches do not perform cross-attention with each other. Finally, the discrete branch predicts categorical logits over the 255 action bins for each discrete action token, while the continuous branch predicts a vector field in the continuous action space. This partially shared design balances the learning of common action features with branch-specific specialization for discrete prediction and continuous refinement.

\subsection{Training}
\looseness=-1
We train the discrete and continuous branches within a paired diffusion framework, while each action chunk $\mathbf{A}_t$ is assigned to only one branch for each training instance.
The discrete branch learns to recover corrupted action tokens, whereas the continuous branch learns a vector field that maps a noisy source initialized from the discrete branch toward the continuous action trajectory.      

We quantize the normalized action chunk $\mathbf{A}_t$ into a discrete sequence $\mathbf{Z}_t = Q(\mathbf{A}_t)$. A subset of discrete action tokens is replaced by the learnable \texttt{[MASK]} token according to the sampled forcing level, producing a corrupted sequence $\mathbf{Z}_t^\rho$, where $\rho\in[0,1]$ denotes the discrete corruption level.
The discrete branch is trained to recover the original action bins only at the masked positions.

Let $\mathcal{M}$ denote the set of masked token indices. The discrete training objective is defined as
\begin{equation}
    \mathcal{L}_{\mathrm{disc}}
    =
    -\frac{1}{|\mathcal{M}|}
    \sum_{i\in\mathcal{M}}
    \log p_{\theta_d}
    \left(
        z_i
        \mid
        \mathbf{Z}_t^\rho,\mathbf{o}_t
    \right),
\end{equation}
where $z_i$ denotes the ground-truth discrete action bin at position $i$, and $p_{\theta_d}$ is the categorical distribution predicted by the discrete branch.

To couple coarse discrete structure with precise continuous generation, we construct the source distribution of the continuous branch from the ground-truth quantized action and Gaussian noise.
we define
\begin{equation}
    \mathbf{S}_t
    =
    (1-\alpha)\boldsymbol{\epsilon}
    +
    \alpha Q^{-1}(\mathbf{Z}_t),
    \qquad
    \boldsymbol{\epsilon}\sim\mathcal{N}(\mathbf{0},\mathbf{I}),
\end{equation}
where $Q^{-1}(\cdot)$ maps discrete action bins back to the normalized continuous action space.
We set $\alpha=0.3$ in experiments.
We then interpolate between this discrete-guided source and the target action chunk:
\vspace{-1mm}
\begin{equation}
    \mathbf{A}_t^\gamma
    =
    \gamma\mathbf{S}_t
    +
    (1-\gamma)\mathbf{A}_t,
    \qquad \gamma\in[0,1], 
\end{equation}
where $\gamma$ denotes the continuous-time variable that parameterizes the flow from the source state to the target action.
The continuous branch predicts the corresponding vector field and is
optimized with
\begin{equation}
    \mathcal{L}_{\mathrm{FM}}
    =
    \mathbb{E}
    \left[
    \left\|
    f_{\theta_c}(\mathbf{A}_t^\gamma,\mathbf{Z}_t,\mathbf{o}_t)
    -
    (\mathbf{S}_t-\mathbf{A}_t)
    \right\|_2^2
    \right].
\end{equation}
To enable one-step continuous denoising at inference time, we further introduce a reconstruction objective that trains the continuous branch to directly recover the target action from the source state.
Given $\mathbf{S}_t$, we obtain
\vspace{-1mm}
\begin{equation}
    \hat{\mathbf{A}}_t^{\mathrm{1step}}
    =
    \mathbf{S}_t
    -
    f_{\theta_c}(\mathbf{S}_t,\mathbf{Z}_t,\mathbf{o}_t),
\end{equation}
and minimize
\begin{equation}
    \mathcal{L}_{\mathrm{1step}}
    =
    \left\|
    \hat{\mathbf{A}}_t^{\mathrm{1step}}
    -
    \mathbf{A}_t
    \right\|_2^2.
\end{equation}
For samples assigned to the continuous branch, the training objective is
\begin{equation}
    \mathcal{L}_{\mathrm{cont}}
    =
    \mathcal{L}_{\mathrm{FM}}
    +
    \lambda_{\mathrm{1step}}\mathcal{L}_{\mathrm{1step}}.
\end{equation}
For samples assigned to the discrete branch, we optimize
\begin{equation}
    \mathcal{L}_{\mathrm{disc}}^{\mathrm{train}}
    =
    \lambda_{\mathrm{disc}}\mathcal{L}_{\mathrm{disc}}.
\end{equation}
We apply $\lambda_{\mathrm{1step}}=0.5$ and $\lambda_{\mathrm{disc}}=0.1$ in experiments.
\subsection{Inference}
At inference time, \method~generates an action chunk with only two forward passes (\textit{i.e.}, 2 function evaluations (NFE)): one discrete prediction followed by one continuous
refinement.
Starting from a fully masked discrete action sequence $\mathbf{Z}_{\mathrm{mask}}$, the discrete branch predicts a coarse action prior in a single forward pass:
\vspace{-2mm}
\begin{equation}
    \hat{\mathbf{Z}}_t
    =
    \arg\max_{\mathbf{Z}}
    p_{\theta_d}
    \left(
        \mathbf{Z}
        \mid
        \mathbf{Z}_{\mathrm{mask}}, \mathbf{o}_t
    \right).
\end{equation}
The argmax is applied independently at each discrete action position. The predicted discrete sequence $\hat{\mathbf{Z}}_t$ is first
dequantized to construct the discrete-guided continuous source:
\begin{equation}
    \hat{\mathbf{S}}_t
    =
    (1-\alpha)\boldsymbol{\epsilon}
    +
    \alpha Q^{-1}(\hat{\mathbf{Z}}_t),
    \qquad
    \boldsymbol{\epsilon}\sim\mathcal{N}(\mathbf{0},\mathbf{I}).
\end{equation}
Starting from $\hat{\mathbf{S}}_t$, which corresponds to $\gamma=1$,
the continuous branch performs a single denoising step:
\vspace{-1mm}
\begin{equation}
    \hat{\mathbf{A}}_t
    =
    \hat{\mathbf{S}}_t
    -
    f_{\theta_c}
    \left(
        \hat{\mathbf{S}}_t,
        \hat{\mathbf{Z}}_t,
        \mathbf{o}_t
    \right).
\end{equation}
The resulting continuous action chunk
$\hat{\mathbf{A}}_t$ is used for robot execution.
\section{Experiments}
\label{sec:experiments}

\begin{table}[t]
\centering
\caption{
\textbf{Comparison on the LIBERO benchmark.}
Prior methods are grouped according to their number of function evaluations (NFE):
iterative generation with more than two NFE, and fast generation with NFE $\leq 2$.
Best results are highlighted in bold.
}
\label{tab:libero_comparison}

\small
\setlength{\tabcolsep}{2.0pt}
\renewcommand{\arraystretch}{1.08}
\resizebox{0.8\textwidth}{!}{
\begin{tabular}{l c c c c c c | c}
\toprule
\textbf{Method}
& \textbf{Parameters}
& \textbf{NFE $\downarrow$}
& \textbf{Spatial}
& \textbf{Object}
& \textbf{Goal}
& \textbf{Long}
& \textbf{Avg.} \\

\midrule
\multicolumn{8}{l}{\textit{Iterative Generation (NFE $>2$)}} \\
\addlinespace[2pt]

DreamVLA~\citep{zhang2026dreamvla}
& 1.6B
& 10
& 97.5 & 94.0 & 89.5 & 89.5 & 92.6 \\

MemoryVLA~\citep{shi2026memoryvla}
& 7B
& 10
& 98.4 & 98.4 & 96.4 & 93.4 & 96.5 \\

$\pi_{0.5}$~\citep{pi05}
& 3B
& 10
& \textbf{98.8} & 98.2 & \textbf{98.0} & 92.4 & 96.9 \\

FlowerVLA~\citep{reuss2025flower}
& 1B
& 8
& 97.5 & 99.1 & 96.1 & 94.9 & 96.9 \\

\midrule
\multicolumn{8}{l}{\textit{Fast Generation (NFE  $\leq 2$)}} \\
\addlinespace[2pt]

MolmoAct-7B-D~\citep{lee2025molmoact}
& 7B
& 1
& 87.0 & 95.4 & 87.6 & 77.2 & 86.6 \\

\addlinespace[2pt]

MIP~\citep{mip}
& 3B
& 2
& 97.6 & 95.8 & 95.2 & 82.2 & 92.7 \\

$\pi_{0.5}$~\citep{pi05}
& 3B
& 2
& 97.2 & 93.6 & \textbf{98.0} & 90.4 & 94.8 \\

CF-VLA~\citep{cfvla}
& 3B
& 2
& 98.0 & 99.2 & 96.6 & 92.0 & 96.5 \\

StarVLA~\citep{community2026starvla}
& 1.5B
& 2
& 96.2 & 98.0 & 95.4 & 90.8 & 95.1 \\

\textbf{Discrete Forcing (Ours)}
& 1.5B
& 2
& \textbf{98.8}
& \textbf{99.4}
& 97.0
& \textbf{95.2}
& \textbf{97.6} \\

\bottomrule
\end{tabular}
}
 \vspace{-1em}
\end{table}
\vspace{-0.8em}
\subsection{Simulation Experiments}

\paragraph{Experiment Settings.}
We evaluate \method~on three popular simulation benchmarks: LIBERO~\citep{liu2023libero}, 
RoboTwin 2.0~\citep{chen2025robotwin}, and RoboCasa-GR1~\citep{nasiriany2024robocasa}, covering single-arm, 
dual-arm, and dexterous manipulation settings.
We adopt StarVLA as the baseline across all benchmarks and follow its official training and evaluation protocols for a fair comparison.
\method~and StarVLA use matched model capacities, with approximately 1.5B parameters on LIBERO and 5.5B parameters on RoboTwin 2.0 and RoboCasa-GR1.
Detailed benchmark descriptions are provided in the Appendix~\ref{app:bench_details}.

On LIBERO, we compare \method~with prior methods under different inference budgets, 
grouping them into iterative generation (NFE $>2$) and fast generation 
(NFE $\leq 2$), as shown in Table~\ref{tab:libero_comparison}.
Among fast-generation methods, \method~achieves the best average success rate of 
97.6\%, outperforming StarVLA by 2.5 percentage points under the same 
1.5B parameter scale and two-NFE inference budget.
Moreover, its overall average exceeds all iterative-generation methods listed 
in the table, despite requiring only two function evaluations.

We further evaluate \method~on RoboTwin 2.0 and RoboCasa-GR1 in Table~\ref{tab:robo_results}. 
Detailed task-level results are provided in Appendices~\ref{app:RoboTwin} and~\ref{app:RoboCasa-GR1}.
On RoboTwin 2.0, \method~achieves the highest average success rate of 59.32\%, improving over StarVLA from 49.80\% to 59.32\%.
Notably, this performance is achieved without any robot-specific pre-training, demonstrating the strong effectiveness and generalization capability of our approach.
Considering the training data of RoboTwin 2.0 is only 50 demonstrations per task, the results further showcase the few-shot learning capabilities.
On RoboCasa-GR1, \method~achieves an average success rate of 56.8\%, outperforming StarVLA by 12.9\% and achieving the best performance among the compared methods.
These results demonstrate that \method~maintains substantial performance gains in more complex robot embodiments and higher-dimensional action spaces.

\begin{table}[t]
    \centering
    \caption{
        \textbf{Mean task success rate (\%) on RoboTwin 2.0 and RoboCasa-GR1.}
    }
    \label{tab:robo_results}
    \hspace*{-0.1\columnwidth}
    \begin{subtable}[t]{0.6\columnwidth}
    \centering
    \caption{RoboTwin 2.0}
    \label{tab:robotwin}
    \small
    \setlength{\tabcolsep}{5pt}
    \begin{tabular}{@{}lc@{}}
        \toprule
        \textbf{Method} & \textbf{Avg.} \\
        \midrule

        DP~\citep{chi2025diffusion}
        & 28.04 \\

        DP3~\citep{ze20243d}
        & 55.24 \\

        RDT~\citep{liu2024rdt}
        & 34.50 \\

        H-RDT~\citep{bi2026h}
        & 37.20 \\        

        TwinVLA~\citep{im2026twinvla}
        & 42.00 \\

        $\pi_{0}$~\citep{black2024pi0}
        & 46.42 \\

        UP-VLA~\citep{zhang2025up}
        & 52.92 \\

        \midrule

        StarVLA~\citep{community2026starvla}
        & 49.80 \\

        \textbf{Discrete Forcing (Ours)}
        & \textbf{59.32} \\

        \bottomrule
    \end{tabular}
\end{subtable}
    \hspace{-0.03\columnwidth}
    \begin{subtable}[t]{0.38\columnwidth}
        \centering
        \caption{RoboCasa-GR1}
        \label{tab:robocasa_gr1}
        \small
        \setlength{\tabcolsep}{3pt}
        \begin{tabular}{@{}lc@{}}
            \toprule
            \textbf{Method} & \textbf{Avg.} \\
            \midrule
            $\pi_{0.5}$~\citep{pi05} & 37.0 \\
            GR00T-N1.6~\citep{gr00tn1_2025} & 47.6 \\
            DiT4DiT~\citep{ma2026dit4dit} & 50.8 \\
            LangForce~\citep{lian2026langforce} & 52.6 \\
            FLARE~\citep{zheng2025flare} & 55.0 \\
            DualCoT-VLA~\citep{zhong2026dualcot} & 55.1 \\
            LDA~\citep{lyu2026lda} & 55.4 \\

            \midrule

            StarVLA~\citep{community2026starvla} & 43.9 \\
            \textbf{\method~(Ours)}
            & \textbf{56.8} \\

            \bottomrule
        \end{tabular}
    \end{subtable}
\end{table}
\vspace{-0.8em}
\subsection{Analysis of Model Scaling}

\begin{figure}[t]
    \centering
    \includegraphics[width=0.6\linewidth]{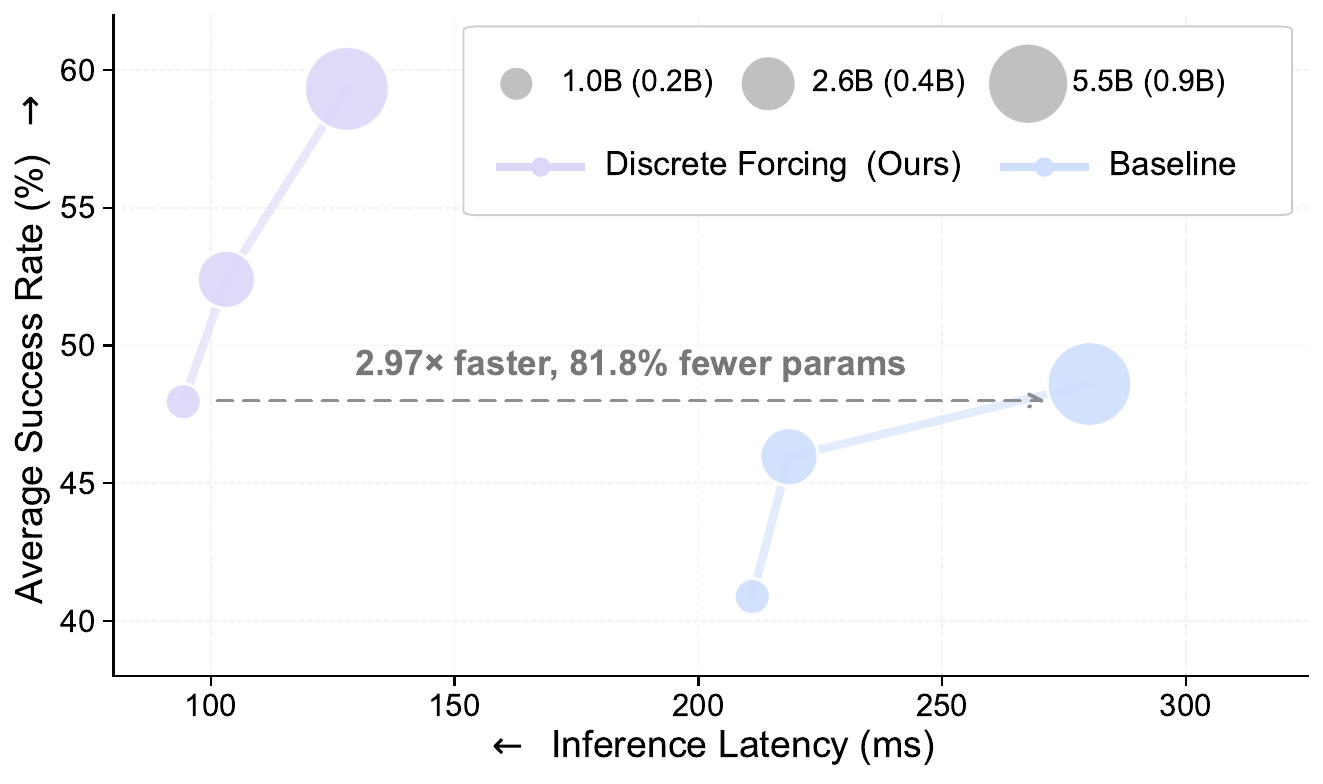}
    \caption{
        \textbf{Model scaling experiments.}
        Marker size denotes model parameter count, while the value in parentheses indicates the parameter count of the action expert.
        Our \method~consistently improves with model scale while
        maintaining lower inference latency than the baseline.
    }
    \label{fig:scaling}
\end{figure}

An ideal action expert should exhibit favorable scaling behavior, with performance improving as its parameter count increases. 
To evaluate the scalability of \method, we train both the StarVLA baseline and our \method~on three model scales under the same training budget on RoboTwin 2.0: 1.0B, 2.6B, and 5.5B parameters, with action experts comprising 175M, 400M, and 900M parameters, respectively.
As shown in Fig.~\ref{fig:scaling}, the performance of \method~improves monotonically with model size, with no evident saturation within the evaluated scale range.
This demonstrates the scalability of our method, indicating that increased model capacity enables it to learn a broader range of tasks and acquire more diverse capabilities.

Moreover, \method~consistently achieves higher success rates and lower inference latency than the corresponding baseline on the same model scale.
In particular, the 1.0B variant of \method~achieves 47.96\% success, approaching the 48.60\% performance of the 5.5B baseline, while being $2.97\times$ faster and using 81.8\% fewer parameters.

\subsection{Inference Efficiency}

\begin{table}[t]
    \centering
    \caption{
    \textbf{Inference efficiency comparison on LIBERO.}
    Action latency measures the action-generation module only, while End-to-End Latency additionally includes the VLM forward pass.
    P50 and P95 denote the median and 95th-percentile latency, respectively. 
    }
    \label{tab:inference_efficiency}

    \resizebox{0.9\linewidth}{!}{
    \begin{tabular}{l cc cc cc}
        \toprule

        \multirow[c]{2}{*}{
            \raisebox{-0.8ex}{\textbf{Method}}
        }
        &
        \multirow[c]{2}{*}{
            \raisebox{-0.8ex}{\textbf{Success Rate (\%)} $\uparrow$}
        }
        &
        \multirow[c]{2}{*}{
            \raisebox{-0.8ex}{\textbf{NFE} $\downarrow$}
        }
        &
        \multicolumn{2}{c}{
            \textbf{Action Latency (ms)} $\downarrow$
        }
        &
        \multicolumn{2}{c}{
            \textbf{End-to-End Latency (ms)} $\downarrow$
        }
        \\

        \cmidrule(lr){4-5}
        \cmidrule(lr){6-7}

        & &
        & \textbf{P50}
        & \textbf{P95}
        & \textbf{P50}
        & \textbf{P95}
        \\

        \midrule

        StarVLA (NFE=10)
        & 96.7
        & 10
        & 145.46
        & 149.36
        & 200.43
        & 206.06 \\

        StarVLA (NFE=4)
        & 96.1
        & 4
        & 58.05
        & 59.88
        & 119.86
        & 123.57 \\

        StarVLA (NFE=2)
        & 95.1
        & 2
        & 29.20
        & 30.44
        & 93.29
        & 96.26 \\

        \midrule

        \textbf{\method~(Ours)}
        & \textbf{97.6}
        & 2
        & \textbf{24.91}
        & \textbf{29.69}
        & \textbf{78.95}
        & \textbf{83.89} \\

        \bottomrule
    \end{tabular}
    }
     \vspace{-0.8em}
\end{table}
To quantitatively evaluate the inference efficiency of \method, we conduct a detailed latency analysis.
As shown in \Cref{tab:inference_efficiency}, reducing the number of denoising steps of StarVLA from 10 to 2 improves inference speed, but also leads to an obvious drop in success rate, revealing a clear trade-off between efficiency and performance.
In contrast, our \method~achieves a substantially higher success rate than the 2-step baseline while retaining the speed benefits of two-step denoising, owing to the structural advantages of its coarse-to-fine denoising scheme.
Despite adopting a dual-branch architecture, \method~maintains the same parameter count and peak memory usage as the standard action expert, with all variants requiring 3.09~GiB of peak GPU memory under our evaluation setting.
Compared with the 10-step baseline, our \method~achieves 5$\times$ acceleration on action generation and more than 2$\times$ acceleration during the complete end-to-end process, with an even higher success rate.
\vspace{-4mm}
\subsection{Mechanism of Structural Action Generation}

\begin{figure}[t]
    \centering
    \includegraphics[width=1\linewidth]{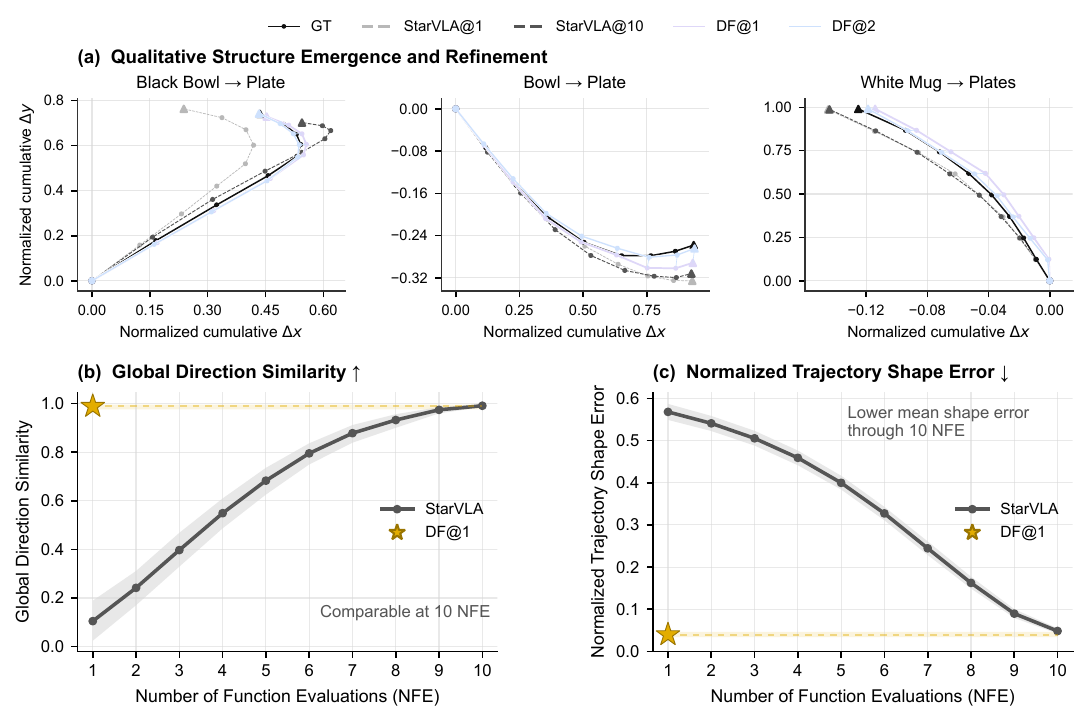}
    \caption{\textbf{Comparison of the action generation mechanisms between our method and the continuous-generation baseline.} DF denotes \method. DF@1 denotes the output after the first discrete forward pass, whereas DF@2 denotes the output after the subsequent continuous refinement. StarVLA@$n$ denotes the action output after the $n$-th iterative forward pass.}
    \label{structure}
     \vspace{-1em}
\end{figure}

To better understand why \method~outperforms the continuous-generation baseline, we investigate the internal action-generation process. We hypothesize that the first discrete prediction of \method~facilitates the early formation of global action structure, compared with iterative continuous generation.

\looseness=-1
We begin with an examination of several action chunks.
For visualization, we convert the translational actions into cumulative relative trajectories. Given translational increments $\Delta \mathbf{p}_h \in \mathbb{R}^3$, $h=1,\ldots,H$, we define
$\mathbf{p}_h = \sum_{k=1}^{h} \Delta\mathbf{p}_k.$
For visualization, each trajectory is further normalized by its path length to reduce the effect of motion magnitude.
As shown in Fig.~\ref{structure}(a), DF@1 already exhibits a trajectory geometry close to the overall motion
pattern of the ground truth,  whereas StarVLA@1 remains less structured. DF@2 further introduces local adjustments around the discrete trajectory.

To further quantify this observation, we introduce two complementary trajectory-level metrics.
\textbf{Global Direction Similarity} evaluates whether the predicted action chunk moves in the correct overall direction, while
\textbf{Normalized Trajectory Shape Error} evaluates how well its trajectory geometry matches the ground truth after removing the effect of motion magnitude.
Higher direction similarity and lower shape error indicate better global action structure.
Detailed metric definitions and analysis settings are provided in Appendix~\ref{app:trajectory_metrics}.

As shown in Fig.~\ref{structure}(b), DF@1 achieves a Global Direction Similarity of $0.989$ with only one NFE. 
StarVLA progressively improves its direction consistency with additional NFEs and reaches a comparable level only at 10 NFEs.
This indicates that the discrete prediction captures the global motion direction at an early generation stage.
Fig.~\ref{structure}(c) further shows that this early structural advantage extends beyond the motion direction.
DF@1 achieves a Normalized Trajectory Shape Error of $0.0399$, which remains lower than that of StarVLA throughout the evaluated 1--10 NFE range.
This suggests that the advantage of the discrete prediction is not limited to the overall motion direction, but also extends to the global geometry of the action trajectory.
These results support our hypothesis that the discrete stage facilitates the early formation of structured action representations, providing a useful structural prior for subsequent continuous refinement.

\subsection{Ablation Studies}
We ablate the key design choices of Discrete Forcing on LIBERO-Long, as summarized in Table~\ref{tab:ablation}.
The generation-order ablation shows that Discrete $\rightarrow$ Continuous achieves the best performance, supporting our coarse-to-fine formulation in which discrete prediction establishes action structure before continuous refinement. In contrast, using the same representation for both stages consistently degrades performance. Introducing explicit branch communication further reduces success rates, favoring independently specialized branches. Removing the discrete-guided source or the one-step training objective also hurts performance. Finally, replacing the ground-truth discrete source during training with the predicted discrete source yields nearly identical performance (95.3\% vs.\ 95.2\%), indicating that the continuous refiner is robust to errors in discrete prediction. We therefore use the ground-truth discrete source during training, which avoids an additional discrete forward pass without sacrificing performance.
Additional ablations on architectural and training configurations are provided in Appendix~\ref{app:ablation}.

\subsection{Real-World Experiments}
\begin{figure}[t]
    \centering
    \includegraphics[width=\linewidth]{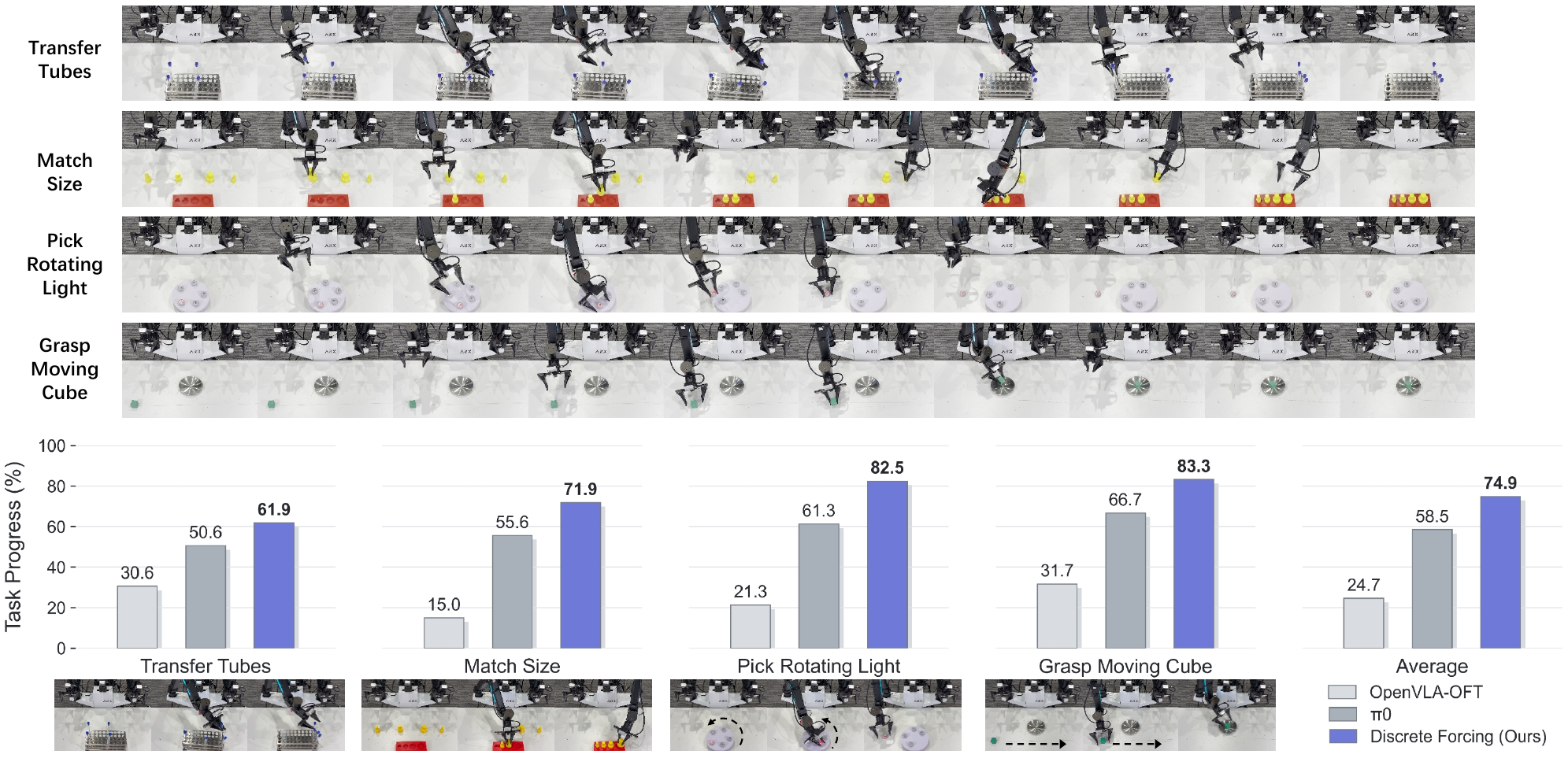}
    \caption{\textbf{Real-world experiments across four tasks.}}
    \label{fig:real}
    \vspace{-1em}
\end{figure}

\begin{table}[t]
    \centering
    \caption{\textbf{Ablation studies on LIBERO-Long.}
    We analyze the effect of generation order, architectural designs,
    and discrete conditioning strategies.}
    \label{tab:ablation}

    \resizebox{0.6\linewidth}{!}{
    \begin{tabular}{l c c}
        \toprule
        \textbf{Variant} & \textbf{Success Rate (\%)} & $\boldsymbol{\Delta}$ \\
        \midrule

        \textbf{\method~(Ours)} & \textbf{95.2} & -- \\

        \midrule
        \multicolumn{3}{c}{\textit{Generation Order}} \\
        \midrule
        $Discrete\!\rightarrow\!Discrete$       & 93.6 & $-1.6$ \\
        $Continuous\!\rightarrow\!Continuous$   & 91.8 & $-3.4$ \\

        \midrule
        \multicolumn{3}{c}{\textit{Architecture and Training Design}} \\
        \midrule
        w/o Discrete-guided Source              & 92.6 & $-2.6$ \\
        w/o One-step Loss                       & 93.4 & $-1.8$ \\
        w/ Branch communication                 & 90.4 & $-4.8$ \\        
        \makecell[l]{w/ Predicted discrete source (train)\\
        \textit{(one extra forward)}}           & 95.3 & $+0.1$ \\

        \bottomrule
    \end{tabular}
    }
\end{table}
All real-world experiments are conducted using an ARX LIFT2 dual-arm robot equipped with three Intel RealSense D405 cameras: one head-mounted camera and one camera mounted on each arm. 
To evaluate our method under real-world manipulation challenges, we design four tasks that cover two complementary capabilities: precision manipulation and dynamic visuomotor response. \textbf{For precise tasks}, \textit{Transfer Tubes} requires inserting 4 tubes into sequential slots, while \textit{Match Size} requires matching 4 cylinders with slots of corresponding shapes and performing accurate insertion. \textbf{For dynamic tasks}, \textit{Pick Rotating Light} evaluates the ability to track and grasp a continuously moving target. \textit{Grasp Moving Cube} requires intercepting and grasping a laterally moving cube. To precisely measure task completion, we report task progress rather than binary success rate. We compare against OpenVLA-OFT~\citep{kim2025fine}, a representative parallel continuous regression VLA, and $\pi_0$~\citep{black2024pi0}, a representative continuous flow-based policy. 
To demonstrate the versatility of our method and avoid potential limitations of evaluating real-world performance with an unpretrained StarVLA backbone, we instantiate our method on top of $\pi_0$ for real-world evaluation.
We train a single model on trajectories collected from all real-world tasks, rather than training a separate model for each task.
Each task is evaluated over 20 independent trials. 
Detailed task settings and the definition of Task Progress are provided in Appendix~\ref{app:real-world}.

As shown in Fig.~\ref{fig:real}, our method consistently outperforms both baselines across all four tasks, achieving an average Task Progress of 74.9\%, corresponding to an improvement of 16.4\%. The gains are particularly pronounced on Pick Rotating Light and Grasp Moving Cube, where successful execution depends critically on rapid visual grounding, dynamic target tracking, and timely action execution. These results indicate that our method improves not only precise manipulation but also responsiveness in latency-sensitive dynamic scenarios.

\section{Conclusion}
We presented \method, a coarse-to-fine action generation framework that unifies discrete and continuous representations within a single flow-matching action expert. By first generating discrete tokens that capture coarse action structure and then using them to initialize continuous denoising, our method combines compact high-level representations with precise action generation. Shared DiT layers and specialized branches enable this process while matching the parameter count of a conventional single-branch DiT, requiring only one forward pass per branch. Extensive evaluations demonstrate improved performance and faster inference over continuous action experts with comparable parameter counts, alongside consistent gains with increasing model capacity. Real-world experiments further validate its effectiveness on high-precision and dynamic manipulation tasks. These findings highlight discrete-to-continuous generation as a promising approach to efficient and precise robot action modeling.

\bibliography{df_paper}
\bibliographystyle{assets/plainnat}

\clearpage
\beginappendix
\section{Implementation Details}
\label{app:impl}
Table~\ref{tab:libero_implementation} summarizes the implementation details and training configurations used for our LIBERO experiments, including model architecture, optimization settings, objective weights, sampling strategies, and evaluation protocol.

\begin{table}[htbp]
    \centering
    \caption{\textbf{Implementation details on LIBERO.}}
    \label{tab:libero_implementation}
    \small
    \setlength{\tabcolsep}{4pt}
    \renewcommand{\arraystretch}{1.08}

    \begin{tabularx}{0.8\columnwidth}{
    >{\centering\arraybackslash}m{0.32\columnwidth}
    >{\centering\arraybackslash}X
}
        \toprule
        \textbf{Configuration} & \textbf{Setting} \\
        \midrule

        Model parameters
        & 1.55B total \\

        Action DiT
        & 24 layers per execution path, with 2 shared and 22 branch-specific layers; hidden dim.\ 1024 \\

        Action horizon
        & 8 \\

        Optimizer
        & AdamW, $\beta=(0.9,0.95)$, $\epsilon=10^{-8}$, weight decay $10^{-8}$ \\

        Learning rate
        & $1\times10^{-5}$ for both VLM and Action Expert \\

        Batch size
        & 16 per GPU on 4 H100 GPUs; global batch size 64 \\

        Training steps
        & 80K optimizer steps \\

        Discrete objective
        & $\lambda_{\mathrm{disc}}=0.1$ \\

        Discrete corruption
        & Beta$(1.5,1.0)$-based schedule; mask positions are sampled uniformly without replacement \\

        Continuous time sampling
        & Beta$(1.5,1.0)$-based paired time schedule with $s=0.999$ \\

        Evaluation
        & 50 rollouts per task; replanning every 8 executed actions \\

        Reported average
        & Arithmetic mean over the four LIBERO suite success rates \\

        \bottomrule
    \end{tabularx}
\end{table}

\FloatBarrier

\section{Detailed Simulation Experimental Settings}
\label{app:bench_details}
We evaluate our method on three simulation benchmarks: LIBERO~\citep{liu2023libero}, RoboTwin 2.0~\citep{ chen2025robotwin}, and RoboCasa-GR1~\citep{nasiriany2024robocasa}.

LIBERO consists of four suites, namely LIBERO-Spatial, LIBERO-Object, LIBERO-Goal, and LIBERO-Long.
It employs a Franka single-arm manipulator and evaluates generalization across different spatial layouts, objects, task goals, and long-horizon manipulation tasks.

RoboTwin 2.0 is a large-scale benchmark for bimanual robotic manipulation, comprising 50 diverse dual-arm tasks that cover a broad range of interaction patterns, including object placement, tool use, stacking, handover, articulated-object manipulation, and coordinated bimanual behaviors. 

RoboCasa-GR1 consists of 24 tabletop manipulation tasks performed by a GR-1 humanoid robot.
It complements the single-arm and dual-arm benchmarks above by evaluating generalist manipulation on a humanoid embodiment across diverse household scenarios.

\section{RoboTwin 2.0 Task-Level Results}

\label{app:RoboTwin}
While the main text reports the aggregate performance on RoboTwin~2.0, Table~\ref{tab:robotwin-task-results} provides the corresponding task-level success rates for a more fine-grained evaluation. We evaluate all 50 tasks following the benchmark protocol, with 50 evaluation episodes conducted for each task. The reported average is computed across all task-level success rates.

\begin{table}[p]
\centering
\small
\caption{\textbf{Task-level success rates (\%) on RoboTwin~2.0.} We report the
success rate for each task, together with the average performance across all tasks.}
\label{tab:robotwin-task-results}
\resizebox{0.8\textwidth}{!}{%
\begin{tabular}{lr@{\hspace{2.5em}}lr}
\toprule
Task & Success Rate & Task & Success Rate \\
\midrule

\texttt{adjust\_bottle} & 100 &
\texttt{place\_can\_basket} & 56 \\

\texttt{beat\_block\_hammer} & 78 &
\texttt{place\_cans\_plasticbox} & 32 \\

\texttt{blocks\_ranking\_rgb} & 26 &
\texttt{place\_container\_plate} & 98 \\

\texttt{blocks\_ranking\_size} & 20 &
\texttt{place\_dual\_shoes} & 60 \\

\texttt{click\_alarmclock} & 64 &
\texttt{place\_empty\_cup} & 96 \\

\texttt{click\_bell} & 60 &
\texttt{place\_fan} & 34 \\

\texttt{dump\_bin\_bigbin} & 80 &
\texttt{place\_mouse\_pad} & 32 \\

\texttt{grab\_roller} & 94 &
\texttt{place\_object\_basket} & 76 \\

\texttt{handover\_block} & 42 &
\texttt{place\_object\_scale} & 68 \\

\texttt{handover\_mic} & 90 &
\texttt{place\_object\_stand} & 72 \\

\texttt{hanging\_mug} & 24 &
\texttt{place\_phone\_stand} & 68 \\

\texttt{lift\_pot} & 50 &
\texttt{place\_shoe} & 80 \\

\texttt{move\_can\_pot} & 40 &
\texttt{press\_stapler} & 92 \\

\texttt{move\_pillbottle\_pad} & 62 &
\texttt{put\_bottles\_dustbin} & 28 \\

\texttt{move\_playingcard\_away} & 92 &
\texttt{put\_object\_cabinet} & 42 \\

\texttt{move\_stapler\_pad} & 18 &
\texttt{rotate\_qrcode} & 80 \\

\texttt{open\_laptop} & 82 &
\texttt{scan\_object} & 56 \\

\texttt{open\_microwave} & 14 &
\texttt{shake\_bottle\_horizontally} & 96 \\

\texttt{pick\_diverse\_bottles} & 34 &
\texttt{shake\_bottle} & 100 \\

\texttt{pick\_dual\_bottles} & 62 &
\texttt{stack\_blocks\_three} & 12 \\

\texttt{place\_a2b\_left} & 44 &
\texttt{stack\_blocks\_two} & 54 \\

\texttt{place\_a2b\_right} & 46 &
\texttt{stack\_bowls\_three} & 42 \\

\texttt{place\_bread\_basket} & 68 &
\texttt{stack\_bowls\_two} & 80 \\

\texttt{place\_bread\_skillet} & 62 &
\texttt{stamp\_seal} & 24 \\

\texttt{place\_burger\_fries} & 90 &
\texttt{turn\_switch} & 46 \\

\midrule
\multicolumn{1}{l}{\textbf{Avg.}}
& \textbf{59.32}
& &
\\
\bottomrule
\end{tabular}%
}
\end{table}

\FloatBarrier
\section{RoboCasa-GR1 Task-Level Results}
\label{app:RoboCasa-GR1}

The main text reports the average success rates on RoboCasa-GR1 for
comparison across different methods.
For a more fine-grained evaluation, Table~\ref{tab:robocasa_task}
provides the per-task success rates for methods whose task-level results
are available.
Following the standard evaluation protocol, we evaluate each of the 24 RoboCasa-GR1 tasks over 50 episodes.
Although $\pi_0.5$ is included in the aggregate comparison in the main text,
we omit it from this table because StarVLA~\citep{community2026starvla} reports only its overall performance on RoboCasa-GR1 without a per-task breakdown.
\begin{table}[htbp]
\centering
\caption{
\textbf{Per-task success rate (\%) on the RoboCasa-GR1 benchmark.}
}
\label{tab:robocasa_task}
\resizebox{\linewidth}{!}{
\begin{tabular}{l|ccccccc}
\toprule
\textbf{Task}
& \textbf{DiT4DiT}
& \textbf{GR00T-N1.6}
& \textbf{StarVLA-$\pi$}
& \textbf{LangForce}
& \textbf{DualCoT-VLA}
& \textbf{LDA}
& \textbf{Ours} \\
\midrule

BottleToCabinetClose
& 48.0 & 51.5 & 26.0 & 72.0 & 66.0 & 76.0 & 74.0 \\

CanToDrawerClose
& 74.0 & 13.0 & 62.0 & 78.0 & 64.0 & 71.0 & 82.0 \\

CupToDrawerClose
& 52.0 & 8.5 & 42.0 & 46.0 & 46.0 & 41.0 & 20.0 \\

MilkToMicrowaveClose
& 50.0 & 14.0 & 50.0 & 56.0 & 58.0 & 52.0 & 62.0 \\

PotatoToMicrowaveClose
& 36.0 & 41.5 & 42.0 & 36.0 & 30.0 & 41.0 & 40.0 \\

WineToCabinetClose
& 42.0 & 16.5 & 32.0 & 46.0 & 38.0 & 57.0 & 56.0 \\

\midrule

FromCuttingboardToBasket
& 52.0 & 58.0 & 40.0 & 66.0 & 44.0 & 65.0 & 44.0 \\

FromCuttingboardToCardboardbox
& 48.0 & 46.5 & 46.0 & 40.0 & 54.0 & 69.0 & 56.0 \\

FromCuttingboardToPan
& 76.0 & 68.5 & 70.0 & 68.0 & 80.0 & 75.0 & 78.0 \\

FromCuttingboardToPot
& 62.0 & 65.0 & 40.0 & 48.0 & 64.0 & 61.0 & 62.0 \\

FromCuttingboardToTieredbasket
& 50.0 & 46.5 & 44.0 & 44.0 & 46.0 & 51.0 & 46.0 \\

\midrule

FromPlacematToBasket
& 50.0 & 58.5 & 44.0 & 54.0 & 48.0 & 53.0 & 54.0 \\

FromPlacematToBowl
& 56.0 & 57.5 & 52.0 & 62.0 & 58.0 & 55.0 & 64.0 \\

FromPlacematToPlate
& 32.0 & 63.0 & 50.0 & 52.0 & 74.0 & 59.0 & 74.0 \\

FromPlacematToTieredshelf
& 18.0 & 28.5 & 28.0 & 24.0 & 26.0 & 24.0 & 28.0 \\

\midrule

FromPlateToBowl
& 56.0 & 57.0 & 52.0 & 54.0 & 50.0 & 53.0 & 56.0 \\

FromPlateToCardboardbox
& 58.0 & 43.5 & 40.0 & 48.0 & 56.0 & 43.0 & 58.0 \\

FromPlateToPan
& 68.0 & 51.0 & 36.0 & 54.0 & 70.0 & 55.0 & 68.0 \\

FromPlateToPlate
& 58.0 & 78.7 & 48.0 & 78.0 & 76.0 & 61.0 & 72.0 \\

\midrule

FromTrayToCardboardbox
& 38.0 & 51.5 & 34.0 & 50.0 & 52.0 & 65.0 & 52.0 \\

FromTrayToPlate
& 56.0 & 71.0 & 64.0 & 58.0 & 64.0 & 63.0 & 78.0 \\

FromTrayToPot
& 54.0 & 64.5 & 44.0 & 62.0 & 70.0 & 55.0 & 48.0 \\

FromTrayToTieredbasket
& 46.0 & 57.0 & 50.0 & 44.0 & 60.0 & 51.0 & 56.0 \\

FromTrayToTieredshelf
& 38.0 & 31.5 & 28.0 & 22.0 & 28.0 & 33.0 & 36.0 \\

\midrule
\textbf{Average}
& 50.8
& 47.6
& 43.9
& 52.6
& 55.1
& 55.4
& 56.8 \\

\bottomrule
\end{tabular}
}
\end{table}

\FloatBarrier
\section{Structural Action Analysis Details}
\label{app:trajectory_metrics}

\noindent\textbf{Global Direction Similarity.}
We evaluate the global motion direction of the predicted action chunk. Specifically, we compute the cosine similarity between its net displacement and that of the ground-truth trajectory:
\begin{equation}
    S_{\mathrm{dir}}
    =
    \frac{
    \mathbf{d}^{\mathrm{pred}} \cdot \mathbf{d}^{\mathrm{gt}}
    }{
    \|\mathbf{d}^{\mathrm{pred}}\|_2
    \|\mathbf{d}^{\mathrm{gt}}\|_2 + \epsilon
    },
    \qquad
    \mathbf{d}=\sum_{h=1}^{H}\Delta\mathbf{p}_h .
\end{equation}
A higher value indicates better agreement in the global motion direction.

\noindent\textbf{Normalized Trajectory Shape Error.}
To evaluate trajectory geometry independently of the overall motion magnitude,
we normalize each relative trajectory by its path length:
\begin{equation}
    L = \sum_{h=1}^{H}\|\Delta\mathbf{p}_h\|_2,
    \qquad
    \bar{\mathbf{p}}_h
    =
    \frac{\mathbf{p}_h}{L+\epsilon},
\end{equation}
and compute the average point-wise distance between the predicted and ground-truth normalized trajectories:
\begin{equation}
    E_{\mathrm{shape}}
    =
    \frac{1}{H}
    \sum_{h=1}^{H}
    \left\|
    \bar{\mathbf{p}}_h^{\mathrm{pred}}
    -
    \bar{\mathbf{p}}_h^{\mathrm{gt}}
    \right\|_2 .
\end{equation}
Lower values indicate better preservation of the global trajectory shape.

\paragraph{Evaluation Protocol.}
We conduct the analysis offline on LIBERO training demonstrations, as the goal is to probe the action-generation dynamics rather than evaluate policy generalization. 
We select five task identities spanning LIBERO-Object, LIBERO-Goal, LIBERO-Spatial, and LIBERO-10,
and sample 40 action chunks of horizon $H=8$ from each task using a fixed random
seed, resulting in 200 action chunks in total.

The statistics in Fig.~\ref{structure} (b,c) are computed by pooling all sampled
action chunks. Since each task contributes the same number of samples, all five
tasks receive equal weight. We report the mean and its 95\% percentile bootstrap
confidence interval, estimated from 2,000 bootstrap resamples.

\section{Extra Ablation Results}
\label{app:ablation}
As shown in Table~\ref{tab:shared_layers}, introducing a small number of shared Action DiT layers substantially improves performance, with two shared layers achieving the highest success rate of 95.2\%.
These results suggest that limited early sharing facilitates common action-relevant representation learning, whereas excessive sharing restricts the specialization required by the discrete prediction and continuous refinement branches.
We further ablate the composition of the discrete-guided continuous source. Table~\ref{tab:source_ratio_ablation} shows that $\alpha=0.3$ achieves the best performance among the tested settings.

\begin{table}[H]
    \centering
    \caption{\textbf{Ablation on the number of shared Action DiT layers on LIBERO-Long.} }
    \label{tab:shared_layers}
    \begin{tabular}{c c}
        \toprule
        \textbf{\# Shared Layers} & \textbf{Success Rate (\%)} \\
        \midrule
        0  & 88.2 \\
        \textbf{2}  & \textbf{95.2} \\
        4  & 95.0 \\
        6  & 93.6 \\
        8  & 93.4 \\
        10 & 90.0 \\
        22 & 89.0 \\
        \bottomrule
    \end{tabular}
\end{table}

\begin{table}[H]
    \centering
    \caption{
    \textbf{Ablation on the composition of the discrete-guided continuous source on LIBERO-Long.}
    $\alpha$ controls the contribution of the dequantized discrete action, while $1-\alpha$ controls the Gaussian noise.
    }
    \label{tab:source_ratio_ablation}
    \small
    \setlength{\tabcolsep}{7pt}
    \renewcommand{\arraystretch}{1.05}

    \begin{tabular}{c c c}
        \toprule
        \textbf{Discrete Action ($\alpha$)}
        & \textbf{Gaussian Noise ($1-\alpha$)}
        & \textbf{Success Rate (\%)} \\
        \midrule
        0.3 & 0.7 & \textbf{95.2} \\
        0.5 & 0.5 & 93.2 \\
        0.7 & 0.3 & 91.0 \\
        \bottomrule
    \end{tabular}
\end{table}

\clearpage
\section{Real-World Experimental Details}
\label{app:real-world}
\begin{figure}[H]
    \centering
    \includegraphics[width=0.6\textwidth]{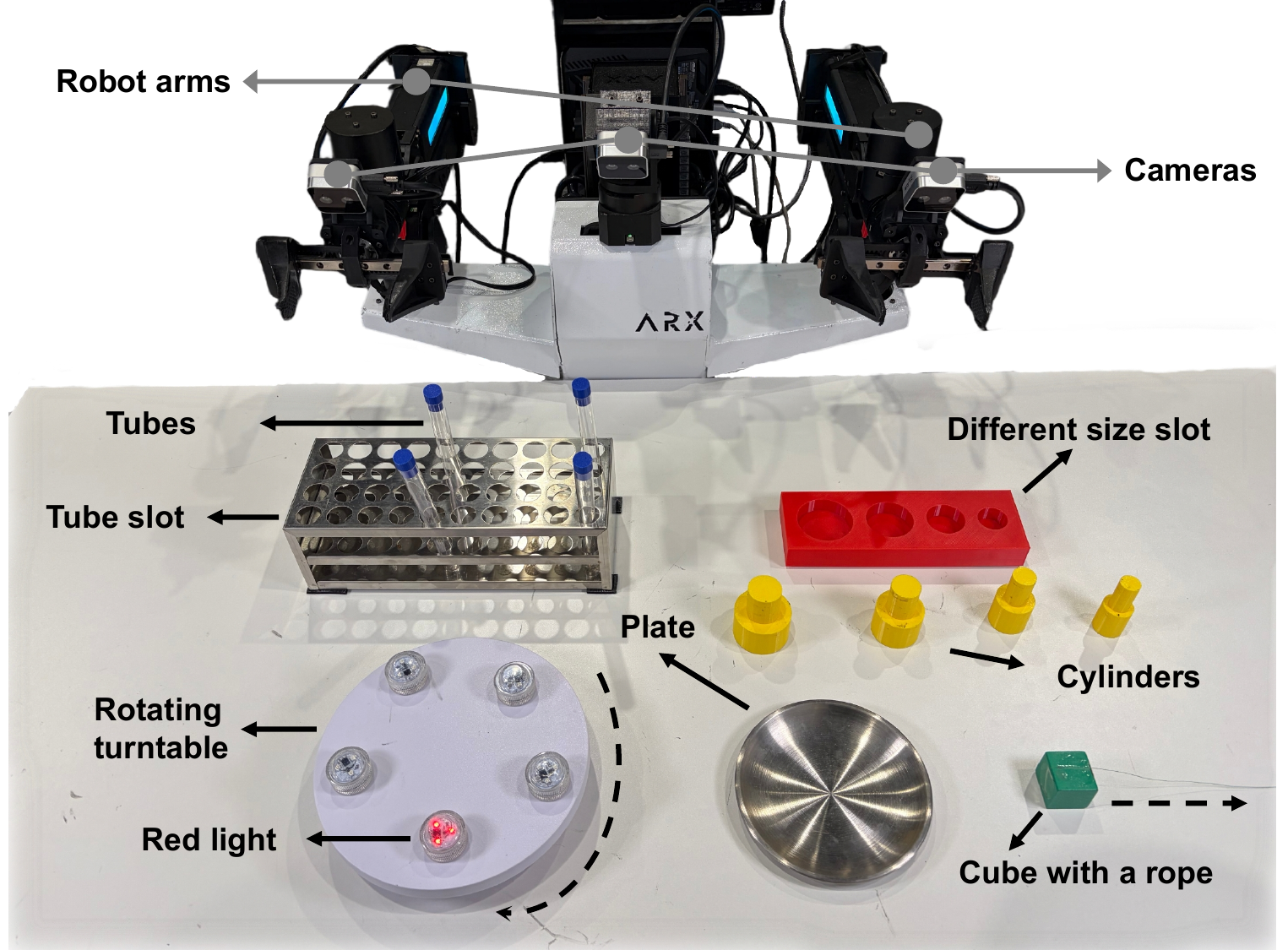}
    \caption{\textbf{Real-world experimental setup.}
    }
\end{figure}

\subsection{Real-World Tasks}
We report the completion rate of each milestone as the task progress rate.

\textbf{Transfer Tubes.}
Four test tubes are initially inserted into randomly selected slots on the right side of a test-tube rack. The robot transfers the tubes one by one to four designated target slots on the left side of the rack. 
The task contains eight milestones:   grasping each test tube and inserting it into the corresponding target slot. 
The task emphasizes repeated fine-grained grasping and precise insertion over a relatively long manipulation sequence. 
The training dataset contains 200 episodes.

\textbf{Match Size.}
Four cylinders of different sizes are randomly positioned in front of a base containing four correspondingly sized holes. The robot must identify the size correspondence between each cylinder and its target hole and insert all four cylinders into the correct locations. 
Each cylinder is grasped by the robot arm that is closer to it after random initialization.
The task contains eight milestones: grasping and inserting each of the four cylinders.
The task requires accurate size-based matching and precise bimanual insertion, as errors in target selection, alignment, or orientation can lead to failure.
The training dataset contains 200 episodes.

\textbf{Pick Rotating Light.}
Five light modules are placed on a continuously rotating turntable, with only one illuminated in red. 
The robot is required to grasp a continuously moving red light and place it on the right side of the workspace.
The task contains four milestones: locating and following the illuminated red light, lifting it from the turntable, moving it away from the rotating platform, and releasing it in the intended right-side area. 
Successful execution requires fast visual grounding and timely manipulation, as delays may cause the moving target to leave the intended grasp position.
The training dataset contains 100 episodes.

\textbf{Grasp Moving Cube.}
A block is initially placed on the right side of the workspace and pulled leftward by a rope. The robot is required to grasp the moving block, and place it onto a designated plate.
The task contains three milestones: locating the moving cube, lifting and transporting it toward the plate, and successfully placing it on the plate. 
As the target remains in motion during the approach, successful execution requires rapid visuomotor response, precise interception, and timely grasping, followed by goal-directed placement.
The training dataset contains 100 episodes.

\end{document}